\documentclass[11pt]{article}
\pdfoutput=1

\usepackage[utf8]{inputenc}
\usepackage[T1]{fontenc}
\usepackage{times} 
\usepackage{amsmath, amssymb, amsfonts}
\usepackage{graphicx}
\usepackage[margin=1in]{geometry}
\usepackage{authblk} 
\usepackage{enumitem}
\usepackage{microtype} 
\usepackage{booktabs}
\usepackage{longtable}
\usepackage{caption}
\usepackage{array}

\usepackage{xcolor}
\usepackage{textcomp}     
\usepackage{xurl}         
\usepackage{listings}     

\usepackage{hyperref}
\hypersetup{
    colorlinks=true,
    linkcolor=blue,
    filecolor=magenta,
    urlcolor=cyan,
    citecolor=blue,
    pdftitle={Deterministic Legal Agents: A Canonical Primitive API for Auditable Reasoning over Temporal Knowledge Graphs},
    pdfauthor={Hudson de Martim},
    pdfkeywords={Legal AI, Legal RAG, SAT-Graph, Temporal Knowledge Graphs, Deterministic Legal Agents, Tool-Augmented Language Models, Neuro-Symbolic AI, Auditability, Explainable AI, Trustworthy AI}
}

\lstdefinestyle{json}{
    basicstyle=\ttfamily\scriptsize, 
    numbers=left,
    numberstyle=\tiny\color{gray},
    stepnumber=1,
    numbersep=5pt,
    showstringspaces=false,
    breaklines=true,
    frame=tb, 
    backgroundcolor=\color{gray!5},
    stringstyle=\color{purple!80!black},
    keywordstyle=\color{blue!70!black},
    commentstyle=\color{green!60!black},
    morestring=[b]",
    escapechar=|, 
    literate=
      {\{}{{{\color{blue!70!black}\lbrace}}}1
      {\}}{{{\color{blue!70!black}\rbrace}}}1
      {[}{{{\color{blue!70!black}[}}}1
      {]}{{{\color{blue!70!black}]}}}1,
}

\lstdefinestyle{tablecode}{
  basicstyle=\small\ttfamily,
  breaklines=true,
  breakatwhitespace=false,
  postbreak=\mbox{\textcolor{red}{\textrightarrow}\space},
  frame=none,
  showstringspaces=false,
  tabsize=2,
  columns=fullflexible, 
  literate=
    *{_}{{\_}}1
     {.}{.}{1}
     {;}{;}{1}
     {:}{:}{1}
     {/}{/}{1}
     {?}{?}{1}
     {=}{=}{1},
}

\title{Deterministic Legal Agents: A Canonical Primitive API for Auditable Reasoning over Temporal Knowledge Graphs}

\author{
Hudson de Martim\\
Federal Senate of Brazil\\
\texttt{hudsonm@senado.leg.br}
}
\date{}

\begin{document}

\maketitle

\begin{abstract}
In high-stakes legal domains, retrieval must do more than identify semantically relevant text: it must preserve the hierarchy, temporality, and causal provenance of legal norms in a reproducible and auditable way. Standard Retrieval-Augmented Generation (RAG), based primarily on semantic similarity over text fragments, cannot reliably provide this level of control. Prior work on SAT-Graph RAG addressed the representation problem by modeling legal materials as structure-aware temporal knowledge graphs. A further question remains: how can an LLM-based reasoning agent interact with such a graph without reintroducing the unreliability that the graph was designed to avoid?

This paper specifies the SAT-Graph API, a canonical primitive interface for auditable reasoning over temporal knowledge graphs, developed and illustrated in the legal domain. The API exposes a library of typed, atomic, and composable primitives that mediate between a probabilistic language model and a deterministic symbolic substrate. Its design is governed by the principle of \textit{Probability Isolation}: stochastic uncertainty is confined to bounded stages of intent translation, initial semantic anchoring, and final narrative synthesis, while structural, temporal, and causal traversals over the graph are executed through deterministic operations over canonical identifiers.

The proposed interface shifts legal RAG from a passive \textit{Retrieve-then-Generate} pipeline to an active \textit{Reason-Act-Observe} process. An agent decomposes a legal question into an explicit execution plan, invokes primitives for point-in-time retrieval, hierarchical context reconstruction, provenance tracing, and impact analysis, and produces an answer grounded in an auditable log of graph operations. The result is not an empirical benchmark, but a formal architectural specification: a secure interaction protocol that decouples legal knowledge representation from agentic reasoning and provides a foundation for trustworthy, explainable, and temporally aware legal AI systems. Although specified and illustrated in the legal domain, the primitive model is intentionally domain-portable: its core abstractions---items, versions, textual units, actions, and relations---can support other temporally versioned, provenance-sensitive, and authority-governed knowledge bases.
\end{abstract}


\section{Introduction}
\label{sec:introduction}

LLM-based agents are increasingly used to reason over external knowledge sources, with Retrieval-Augmented Generation (RAG) emerging as a dominant architecture for grounding their outputs in retrieved evidence~\cite{lewis2020retrieval}. In high-stakes legal domains, however, retrieval cannot be reduced to semantic similarity over text fragments. Legal corpora are structured normative systems: provisions are embedded in hierarchical instruments, evolve across time, and derive their authority and content from identifiable acts of legal change. A provision's meaning therefore depends not only on its wording, but also on its position within a legal instrument, its validity at a given point in time, and the legislative acts that produced, amended, or repealed it. This makes the standard ``flat-text'' formulation of RAG poorly suited to reliable legal reasoning.

Prior work has identified three recurring failure modes of flat-text legal retrieval~\cite{demartim2025graphrag}. First, \textit{mereological blindness}: retrieval over isolated chunks often ignores part-whole relations such as clause $\subset$ paragraph $\subset$ article $\subset$ chapter. Second, \textit{diachronic naivety}: semantically similar provisions from different temporal states may be conflated, causing superseded and currently valid norms to appear interchangeable. Third, \textit{causal opacity}: standard RAG does not preserve the legislative provenance of a provision or the chain of events that produced its current legal state. In legal settings, these failures are not merely technical imperfections. They undermine auditability and can produce outputs that appear fluent while being legally anachronistic or unsupported, a risk made concrete by documented cases of fabricated legal authorities in LLM-assisted litigation~\cite{mata2023avianca,dahl2024largelegalfictions}.

SAT-Graph RAG was introduced to address the representation side of this problem~\cite{demartim2025graphrag}. Instead of treating legal texts as independent chunks, it models legal norms as a structure-aware temporal knowledge graph. The graph represents legal instruments and their components as versionable entities, encodes hierarchical containment, separates textual content from temporal versions, and reifies legislative actions as first-class causal events. This provides a verifiable substrate for legal retrieval: a knowledge base in which structure, validity, applicability, and provenance can be queried explicitly.

A verifiable substrate, however, is not sufficient on its own. A second problem arises at the interaction layer: \textit{how should an LLM-based reasoning agent query such a graph without bypassing its deterministic guarantees?} If the agent is allowed to rely only on unstructured semantic search, the graph's formal structure is effectively discarded at query time. If the agent is allowed to generate arbitrary database queries, such as SQL, SPARQL, or Cypher, the system becomes vulnerable to syntactic confabulation, silent temporal errors, schema coupling, and unsafe query behavior~\cite{scholak2021picard}. In both cases, the uncertainty of the language model leaks into the retrieval process itself.

This paper addresses that interaction problem by specifying the \textbf{SAT-Graph API}: a canonical primitive interface for auditable reasoning over temporal knowledge graphs, developed and illustrated in the legal domain. The API is organized as a library of typed, atomic, and composable operations. Each primitive exposes a controlled form of interaction with the graph: resolving references, retrieving valid versions, reconstructing hierarchical context, tracing provenance, identifying forward impact, searching textual units under structural and temporal constraints, and hydrating retrieved evidence into agent-consumable form. Rather than exposing the physical database schema, the API provides a procedural facade through which an agent can build explicit and inspectable reasoning plans.

The architecture separates three layers. The \textbf{SAT-Graph} is the verifiable legal substrate. The \textbf{Canonical Primitive API} is the deterministic interaction protocol. The \textbf{LLM agent} is the probabilistic planner and narrator. This separation is governed by the principle of \textit{Probability Isolation}. Stochastic uncertainty is confined to three bounded stages: intent translation, where the user query is transformed into an execution plan; initial semantic anchoring, where approximate search may be used to identify candidate graph anchors---including items, themes, item types, or scored text units linked to canonical graph objects---rather than final evidence; and final narrative synthesis, where the verified results are rendered as prose. Once a canonical identifier and temporal reference have been established, structural, temporal, and causal traversals over the graph are deterministic, conditional on the correctness and completeness of the underlying graph state.

The resulting architecture shifts legal RAG from a passive \textit{Retrieve-then-Generate} pipeline to an active \textit{Reason-Act-Observe} process~\cite{yao2023react}. The agent no longer receives a single bundle of retrieved chunks and generates an answer from them. Instead, it decomposes a legal question into a sequence of primitive calls, observes intermediate results, refines its plan when necessary, and produces a final answer grounded in an auditable execution log. This makes the retrieval process itself inspectable: a user can determine which provision was resolved, which temporal version was selected, which action produced it, which structural ancestors contextualized it, and which textual units were ultimately used as evidence.

This paper therefore contributes not another monolithic retrieval system, but a formal interaction protocol for legal agents operating over temporal knowledge graphs instantiated in the legal domain. Its main contributions are:

\begin{itemize}
    \item We specify a \textbf{Canonical Primitive API} for querying structure-aware temporal knowledge graphs, with a primary instantiation in legal norms.
    \item We articulate the \textbf{Probability Isolation} principle, a neuro-symbolic design pattern that confines probabilistic uncertainty to intent translation, semantic anchoring, and narrative synthesis, while preserving deterministic graph traversal after canonical resolution.
    \item We show how the API supports the \textbf{Reason-Act-Observe} paradigm by enabling legal agents to decompose complex questions into explicit, auditable execution plans.
    \item We define a separation of responsibilities between the legal knowledge substrate, the API interaction protocol, and the LLM-based planner, thereby decoupling knowledge representation from agentic reasoning.
    \item We illustrate the API through use cases aligned with the SAT-Graph RAG framework, including point-in-time retrieval, provenance reconstruction, and impact analysis.
\end{itemize}

The remainder of the paper is organized as follows. Section~\ref{sec:related_work} discusses related work in legal information retrieval, Graph RAG, tool-augmented language models, and neuro-symbolic reasoning. Section~\ref{sec:api_specification} presents the architectural foundations, data models, and primitive specification of the SAT-Graph API. Section~\ref{sec:use_cases} illustrates how the primitives operationalize the core SAT-Graph RAG use cases. Section~\ref{sec:conclusion} concludes with limitations, implications, and directions for empirical evaluation.

\section{Related Work}
\label{sec:related_work}

This work lies at the intersection of legal knowledge representation, retrieval-augmented generation, knowledge graph retrieval, and tool-augmented agentic reasoning. We review these areas with a specific focus on the gap addressed by this paper: the lack of a controlled, auditable interaction protocol between an LLM-based reasoning agent and a temporal knowledge graph.

\subsection{Legal AI and Legal Knowledge Representation}

The application of artificial intelligence to legal reasoning has a long history, ranging from rule-based systems and logic programming approaches~\cite{sergot1986logic} to transformer-based models specialized for legal text, such as Legal-BERT~\cite{chalkidis2020legalbert}. Despite these advances, many contemporary retrieval and question-answering systems still treat legal materials primarily as unstructured text. This representation is insufficient for legal corpora whose meaning depends on formal structure, temporal validity, institutional authority, and provenance.

A parallel tradition in legal informatics has addressed this issue through formal models and ontologies. Work on legal resource identification and diachronic evolution has long recognized the need to track legal resources across time~\cite{lima2008ontology}. Our SAT-Graph framework builds on this tradition by modeling legal norms as versionable entities with explicit structural containment and legislative causality~\cite{demartim2025graphrag,demartim2025temporal}. In this paper, we do not propose a new ontology for legal norms. Instead, we specify the interaction layer through which an agent can query such an ontology safely and auditably.

\subsection{Retrieval-Augmented Generation in Legal Domains}

Retrieval-Augmented Generation (RAG) has become a standard architecture for grounding LLM outputs in external sources~\cite{lewis2020retrieval}. In its canonical form, RAG embeds text fragments, retrieves the most similar chunks to a query, and provides those chunks as context for generation. This paradigm is effective for many knowledge-intensive tasks, but it is structurally limited in legal domains.

Legal retrieval requires more than semantic similarity. A correct answer may depend on whether a provision was valid at a specific date, whether it was amended or repealed, which higher-level structural container gives it scope, and which legislative event produced its current wording. Recent legal RAG benchmarks and evaluations highlight that retrieval remains a bottleneck for legal question answering, especially when tasks require temporal or multi-hop reasoning rather than lexical overlap~\cite{pipitone2024legalbench,zheng2025reasoning}. The limitation is therefore not merely that legal RAG needs better embeddings; it needs retrieval mechanisms that expose structure, validity, and provenance as first-class constraints.

\subsection{Knowledge Graphs and Graph RAG}

Knowledge graphs offer a natural response to the limitations of flat-text retrieval. Recent Graph RAG approaches seek to improve retrieval by using graph structure to organize documents, entities, relations, and communities~\cite{pan2024unifying,edge2024graphrag}. These approaches show that graph-based context can support reasoning patterns that are difficult to obtain from isolated text chunks.

However, legal knowledge graphs differ from many general-purpose Graph RAG settings. The relevant graph is not merely an emergent network of entities extracted bottom-up from text. In legal corpora, much of the graph is already normatively defined: statutes contain articles, articles contain paragraphs, amendments create and terminate versions, and legal events produce authoritative changes. SAT-Graph RAG therefore models the legal graph top-down, using the formal structure of legal instruments and the causal structure of legislative actions as the graph's backbone~\cite{demartim2025graphrag}. This paper takes that representation as its substrate and asks a different question: how should an agent interact with it?

\subsection{Tool-Augmented Agents and Neuro-Symbolic Reasoning}

Recent work has shifted LLMs from passive generators to agents capable of planning, acting, and using tools. Tool-Augmented Language Models~\cite{parisi2022talm,schick2023toolformer}, Chain-of-Thought prompting~\cite{wei2022chain}, ReAct-style reasoning~\cite{yao2023react}, and self-refinement methods such as Reflexion~\cite{shinn2023reflexion} provide general patterns for decomposing tasks into intermediate actions. These methods are highly relevant to legal AI because complex legal questions often require multi-step retrieval and verification.

Yet agentic reasoning is only as reliable as the tools exposed to the agent. A legal agent that can issue arbitrary SQL, SPARQL, or Cypher queries may produce syntactically plausible but semantically flawed queries, including queries that omit temporal validity filters or traverse the wrong relation type. Text-to-SQL systems continue to face reliability challenges even in non-legal domains~\cite{pourreza2024dinsql,scholak2021picard}. In legal settings, such failures may be silent: the returned result can be well-formed while being legally anachronistic.

Our proposal follows the neuro-symbolic view that neural models should handle flexible language understanding and planning, while symbolic systems should enforce formal constraints and deterministic execution~\cite{garcez2015neural,besold2021neural}. The SAT-Graph API operationalizes this separation. The LLM agent plans and composes operations; the API validates and executes typed primitives over the graph.

\subsection{Positioning Our Contribution}

Prior work has addressed either the representation problem or the agentic planning problem. SAT-Graph RAG addresses the representation problem by modeling legal norms as temporal, structure-aware, causally traceable knowledge graphs~\cite{demartim2025graphrag}. Tool-augmented LLM research addresses the planning problem by showing how language models can select and sequence external operations. The missing component is the interaction protocol between these two layers.

This paper specifies that protocol as a \textbf{Canonical Primitive API}. The API does not replace the legal knowledge graph, nor does it replace the LLM-based reasoning agent. It mediates between them. Its primitives expose only controlled, typed, and auditable operations: probabilistic discovery operations for initial anchoring, and deterministic graph operations once canonical identifiers have been resolved. This design makes the agent's retrieval process inspectable as an execution plan rather than hidden inside either an embedding search or an unconstrained generated query. The contribution is therefore not a new monolithic retrieval system, but a formal interface for auditable agentic reasoning over temporal knowledge graphs.

\section{Specification of the Canonical Primitive API}
\label{sec:api_specification}

The central thesis of this paper is that a temporal knowledge graph alone is not sufficient for trustworthy\footnote{Throughout this paper, we use \textit{trustworthy} in an operational, retrieval-centered sense: a system is trustworthy to the extent that the evidentiary basis of its answers can be inspected, reproduced, contested, and traced to canonical graph objects. This is narrower than a general claim of legal correctness or autonomous legal judgment.} agentic reasoning. The graph may encode hierarchy, validity, applicability, and provenance with formal precision, but an LLM-based agent can still bypass those guarantees if it interacts with the graph through unconstrained semantic search or generated database queries. What is required is a disciplined interaction layer.

We specify this layer as the \textbf{SAT-Graph API}: a canonical primitive interface through which a reasoning agent queries a temporal knowledge graph using typed, auditable operations. The API acts as a procedural facade over the underlying storage implementation, which may be a graph database, a relational-vector system, or another hybrid architecture. The agent does not need to know the physical schema or generate database-specific query language. It interacts with the knowledge base through a fixed vocabulary of primitives.

The API distinguishes two classes of operations. \textbf{Discovery primitives} operate at the boundary between natural language and the graph. They may use semantic search or entity linking to resolve ambiguous user references into ranked candidate identifiers. \textbf{Deterministic primitives} operate after a canonical identifier has been selected. Given the same identifier, temporal parameters, and graph state, they return the same structural, temporal, textual, or causal result. This distinction is the operational basis of \textit{Probability Isolation}: uncertainty is permitted during anchoring, but graph traversal is deterministic after anchoring.

\begin{figure}[htbp]
    \centering
    \includegraphics[width=0.7\textwidth]{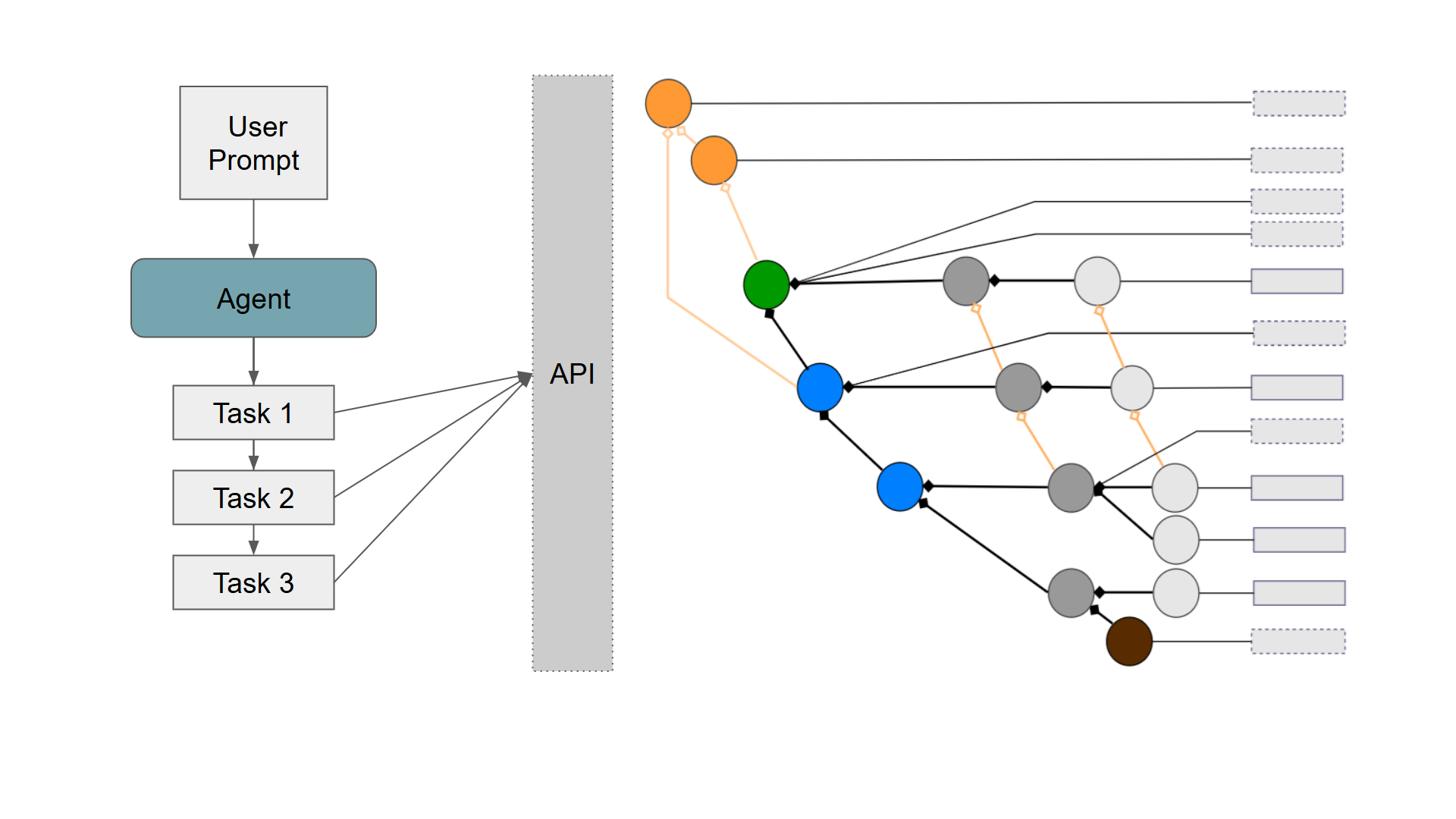}
    \caption{A reasoning agent decomposes a user prompt into tasks executed through primitives provided by the SAT-Graph API.}
    \label{fig:sat_graph_api}
\end{figure}

Figure~\ref{fig:sat_graph_api} illustrates the resulting architecture. The LLM agent decomposes the user's question into an execution plan, invokes primitives, observes intermediate results, and synthesizes a final answer. The SAT-Graph remains the verifiable substrate; the API is the interaction protocol; the agent is the planner and narrator.

Before presenting the data models and primitive groups discussed in this paper, we introduce four architectural foundations: the shift from single-shot retrieval to active agentic reasoning; dynamic context reconstruction; the \textit{Probability Isolation} principle; and the design principles that govern the API surface.

\subsection{From Single-Shot Retrieval to Active Agentic Reasoning}
\label{subsec:paradigm_shift}

In conventional RAG architectures, retrieval follows a \textit{Retrieve-then-Generate} pattern: the system retrieves text fragments by semantic similarity and injects them into the LLM's context. In this model, the LLM is largely a \textbf{passive consumer} of retrieved material. It receives a fixed context window and must generate an answer from whatever the retrieval component delivered, even if the retrieved material lacks the relevant temporal version, structural ancestor, or causal provenance.

The SAT-Graph API supports a different pattern: \textit{Reason-Act-Observe}. In this model, the LLM is an \textbf{active orchestrator}. It interprets the user's question, constructs an execution plan, invokes primitives, observes their outputs, and refines the plan when necessary~\cite{yao2023react}. For example, an agent may first resolve a reference to a legal provision, then retrieve the version valid at a target date, then recover its structural ancestors, then trace the action that produced it, and only then synthesize an answer.

This shift is essential for temporal and authority-governed domains. A legal question is rarely answered by retrieving a single semantically similar paragraph. It may require checking whether the retrieved provision was in force, whether its wording had changed, whether a higher-level title narrows its scope, or whether a later action terminated its validity. The API provides the agent with the primitive operations needed to perform this investigation while keeping each step explicit and auditable.

\subsection{Dynamic Context Reconstruction}
\label{subsec:context_reconstruction}

Granular retrieval improves precision but creates a risk of \textbf{semantic fragmentation}. A retrieved provision may depend on hierarchical context that is not physically adjacent in the text. Legal interpretation often requires knowing not only the wording of a clause or item, but also the article, chapter, title, or instrument in which it is embedded.

Consider Article~5, II, of the 1988 Brazilian Constitution: ``no one shall be obliged to do or refrain from doing something except by virtue of law.'' Read in isolation, this provision may be treated as a generic reference to legality. Its placement within Title~II, Chapter~I---``Individual and Collective Rights and Duties''---together with the wording of Article~5's \textit{caput}, clarifies that the provision functions as a fundamental-rights guarantee: obligations and prohibitions imposed on individuals must have a legal basis.

Flat-text RAG systems usually address this problem with overlapping chunks or larger context windows. These methods rely on physical proximity. They may include adjacent provisions that are irrelevant while still failing to recover distant structural ancestors that determine scope. By contrast, the SAT-Graph API exposes topological proximity. Once an item has been resolved, the agent can explicitly retrieve ancestors, descendants, siblings, or a bounded context tree.

The result is a separation between storage and presentation. The graph stores legal materials granularly, preserving precise identifiers and temporal versions. The agent reconstructs context dynamically, expanding from the retrieved node only when the question requires it. This allows the system to combine the precision of atomic retrieval with the semantic richness of hierarchical interpretation.

\subsection{The Probability Isolation Principle}
\label{subsec:probability_isolation}

The SAT-Graph API is designed around the principle of \textit{Probability Isolation}. The goal is not to eliminate all stochastic behavior from an LLM-based system. Rather, the goal is to locate such behavior in bounded and inspectable stages, while preserving deterministic execution for structural, temporal, and causal graph operations.

In conventional RAG systems, uncertainty permeates the retrieval process. A semantically similar chunk may be retrieved because it is lexically close to the question, even if it belongs to the wrong temporal state, the wrong structural scope, or the wrong legal instrument. In the proposed architecture, probabilistic operations are permitted only where linguistic ambiguity makes them necessary. Once the relevant graph entity has been anchored through a canonical identifier, the remaining operations are symbolic traversals over the graph.

This yields a reasoning flow with three probabilistic stages surrounding a deterministic core.

\paragraph{Stage 1 --- Intent Translation.}
The LLM interprets the user's natural-language query and proposes an execution plan. Errors at this stage may include malformed parameters, non-existent primitive names, invalid dates, or inappropriate sequencing of operations. Because the API exposes a typed interface, many such errors can be detected as validation failures before graph execution. The result is an explicit planning error rather than a silent retrieval error.

\paragraph{Stage 2 --- Initial Semantic Anchoring.}
When the user refers to an entity, asks a thematic question, or describes a legal problem in natural language, the system may need to perform semantic search, lexical search, or entity linking to identify candidate graph anchors. This stage is probabilistic because references such as ``Article~6'', topical queries such as ``the right to strike'', or open-ended questions such as ``under which circumstances does a worker have the right to strike?'' may map to several provisions, themes, text units, or relations. Discovery primitives therefore return ranked candidates with scores and metadata rather than silently selecting a single result. The agent may proceed when confidence is sufficient, combine multiple anchors when the question requires a broader legal regime, or request clarification when ambiguity remains.

\paragraph{Deterministic Core --- Symbolic Graph Execution.}
After a canonical identifier and relevant temporal parameters have been established, the API executes deterministic graph operations. Given the same graph state and the same inputs, primitives such as version retrieval, ancestor reconstruction, provenance tracing, and impact analysis return the same results. This determinism is conditional on the correctness and completeness of the underlying graph, but it removes semantic approximation from the traversal itself.

\paragraph{Stage 3 --- Narrative Synthesis.}
The final answer is produced by the LLM from the retrieved evidence. This stage remains probabilistic and can still introduce paraphrasing errors or overgeneralized conclusions. However, the evidentiary substrate is no longer hidden inside a vector search result. Each factual claim can be checked against the execution log of primitive calls and graph outputs.

Probability Isolation therefore changes the nature of system error. It does not guarantee that the agent will always choose the correct plan, nor does it eliminate the need for human review in high-stakes legal settings. It ensures that errors become more localizable. Anchoring errors are visible in candidate selection; planning errors are visible in the primitive sequence; synthesis errors are visible by comparing the answer to the retrieved evidence.

A residual class of \textbf{semantic planning errors} remains. An agent may invoke syntactically valid primitives that are logically inadequate for the user's question, such as querying the wrong legal instrument or failing to retrieve a relevant related norm. The architecture does not hide this risk. Instead, it makes the plan auditable. In this sense, the API functions as a constrained execution environment for legal reasoning: the agent writes the plan, the API validates and executes typed operations, and the graph provides the deterministic runtime over curated legal data.

\subsection{Design Principles}
\label{subsec:design_principles}

The API is governed by three design principles that operationalize Probability Isolation and support auditable agentic reasoning.

\paragraph{Determinism after Anchoring.}
The API does not claim that every interaction is deterministic. Discovery primitives necessarily operate under uncertainty because they translate natural-language references, thematic descriptions, or semantic queries into candidate graph anchors\footnote{In this paper, an \textit{anchor} is any graph-grounded object that allows the agent to move from probabilistic discovery to deterministic traversal: an entity identifier, a theme, an item type, or a scored \texttt{TextUnit} linked to a canonical source object.}. These anchors may be entity identifiers, such as \texttt{Item}, \texttt{Theme}, or \texttt{ItemType} IDs, or textual anchors, such as scored \texttt{TextUnit} results linked to canonical graph objects through \texttt{sourceType} and \texttt{sourceId}. However, once the agent selects a canonical anchor, or a bounded set of canonical anchors, and the relevant temporal parameters are fixed, primitives that traverse structure, time, text, or causality behave as deterministic read operations over the graph. This distinction prevents probabilistic discovery from contaminating the entire reasoning process.

\paragraph{Composability.}
Primitives are atomic building blocks rather than monolithic reasoning procedures. The API exposes operations such as resolving references, retrieving valid versions, reconstructing hierarchy, tracing provenance, searching text units, and hydrating evidence. The reasoning logic that combines these operations remains the responsibility of the agent. This separation makes the plan explicit: complex legal reasoning is represented as a sequence or graph of primitive calls rather than as an opaque server-side procedure.

\paragraph{Auditability.}
Every primitive call can be logged with its inputs, outputs, timestamps, and confidence or provenance metadata. Probabilistic primitives expose ranked candidates and scores. Deterministic primitives expose canonical identifiers, version intervals, action links, and textual evidence. The resulting execution trace allows a human reviewer to inspect how an answer was grounded: which entity was selected, which version was retrieved, which action produced it, which context was reconstructed, and which text units were used for synthesis.

\subsection{Core Data Models}
\label{subsec:data_models}

The primitives exposed by the SAT-Graph API operate over a small set of canonical data models. These models are intentionally named at a level of abstraction above the legal domain. Although the API is specified and illustrated through legal norms, its core entities---\texttt{Item}, \texttt{Version}, \texttt{TextUnit}, \texttt{Action}, and \texttt{Relation}---can represent other temporally versioned and authority-governed knowledge bases. In the legal instantiation, they correspond to legal instruments, structural components, temporal states, textual realizations, legislative or institutional events, and semantic links.

This abstraction is not a loss of legal precision. Rather, it is the mechanism through which the API decouples domain ontology from agent-facing interaction. The SAT-Graph ontology may distinguish norms, components, temporal versions, language versions, legislative events, and semantic relations~\cite{demartim2025graphrag,demartim2025temporal}; the API exposes these concepts through a compact vocabulary that an LLM-based agent can reliably use in execution plans. Domain-specific distinctions remain available through fields such as \texttt{type}, \texttt{theme}, \texttt{predicate}, and \texttt{metadata}.

\begin{figure}[htbp]
    \centering
    \includegraphics[width=0.69\textwidth]{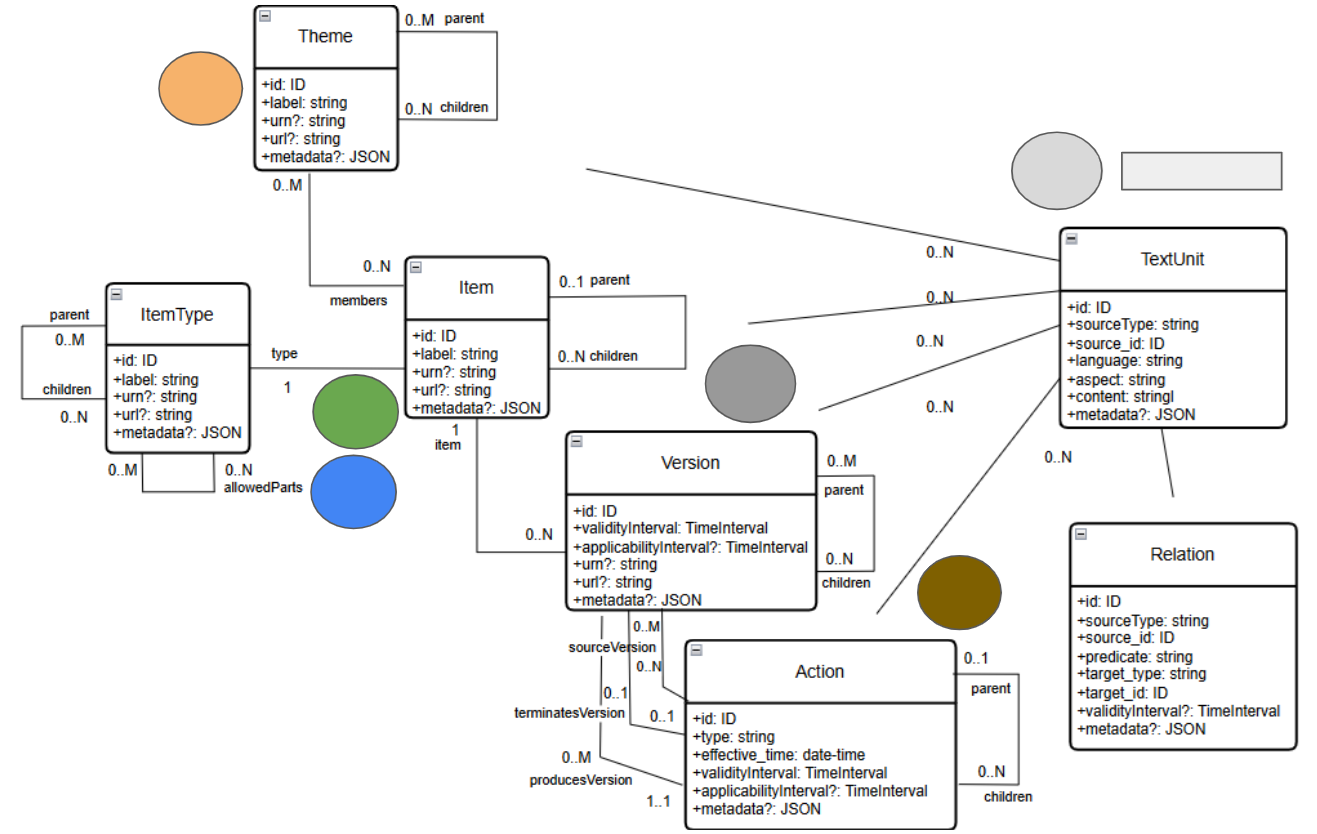}
    \caption{SAT-Graph Ontology as exposed by the API.}
    \label{fig:satgraph_ontology}
\end{figure}

Figure~\ref{fig:satgraph_ontology} illustrates the resulting data model as exposed by the API.

\begin{table}[htbp]
\centering
\caption{Canonical data models exposed by the SAT-Graph API.}
\label{tab:data_models_summary}
\begin{tabular}{p{0.18\textwidth} p{0.33\textwidth} p{0.39\textwidth}}
\toprule
\textbf{Model} & \textbf{Role in the API} & \textbf{Primary agent use} \\
\midrule
\texttt{Item} & A canonical graph entity with stable identity. & Provides stable identity for retrieval, hierarchy, and reference resolution. \\

\texttt{Theme} & Conceptual category in a thematic taxonomy. & Scopes search and supports subject-matter exploration. \\

\texttt{ItemType} & Structural type in the schema-level taxonomy. & Enables schema introspection and type-constrained retrieval. \\

\texttt{Version} & Time-bound state of an \texttt{Item}. & Supports point-in-time retrieval, temporal comparison, and validity reasoning. \\

\texttt{TextUnit} & Textual or semantic realization attached to a graph entity. & Bridges semantic search, evidence extraction, and final answer synthesis. \\

\texttt{Action} & Reified state-transition event that produces, terminates, or modifies \texttt{Version}s. & Supports provenance tracing and forward impact analysis over graph mutations. \\

\texttt{Relation} & First-class semantic edge between graph entities. & Supports citation, implementation, regulation, and other cross-reference traversal. \\
\bottomrule
\end{tabular}
\end{table}

\begin{description}
    \item[\texttt{Item}:] Represents a canonical graph entity with stable identity. In the legal-norm submodel, structural \texttt{Item}s are commonly instantiated as \texttt{Work} or \texttt{Work Component}. Other domain instantiations may introduce additional \texttt{ItemType}s, such as institutions, offices, actors, technical artifacts, or documentary objects, provided they preserve stable identity and are connected to the graph through versions, relations, or metadata as appropriate.
    \begin{itemize}
        \item \texttt{id (ID):} A unique, canonical identifier.
        \item \texttt{typeId (ID):} The specific item type ID (e.g., ``Constitution,'' ``Article,'' ``Paragraph'').
        \item \texttt{label (string):} The human-readable label.
        \item \texttt{urn (string)?:} (Optional) A persistent semantic identifier (e.g., a canonical LEX URN).
        \item \texttt{url (string)?:} (Optional) A dereferenceable URI providing Web access to the object.
        \item \texttt{parentId (ID)?:} (Optional) The ID of its single structural parent. A \textbf{\texttt{Work}} (the root concept) has no parent.
        \item \texttt{metadata (JSON)?:} (Optional) A flexible JSON store for structured properties of the item~\cite{demartim2025structuredRepresentation}.
    \end{itemize}

    \item[\texttt{Theme}:] Represents a conceptual category used to classify \texttt{Item}s and organize knowledge, for example through a SKOS-like taxonomy. Themes form a directed acyclic graph (DAG) to support a poly-hierarchical taxonomy.
    \begin{itemize}
        \item \texttt{id (ID):} A unique, canonical identifier.
        \item \texttt{label (string):} The human-readable (preferred) label.
        \item \texttt{urn (string)?:} (Optional) A persistent semantic identifier.
        \item \texttt{url (string)?:} (Optional) A dereferenceable URI providing Web access.
        \item \texttt{parentIds (list[ID])?:} (Optional) A list of IDs of broader themes. Themes can have multiple parents; root themes have an empty list.
        \item \texttt{childrenIds (list[ID])?:} (Optional) A list of IDs of more specific (narrower) themes.
        \item \texttt{metadata (JSON)?:} (Optional) A flexible JSON store for structured properties of the theme (e.g., synonyms or scope notes for semantic disambiguation).
    \end{itemize}

    \item[\texttt{ItemType}:] Represents the structural meta-model exposed to the agent. It defines both the taxonomy (\textit{Is-A}) and the composition rules (\textit{Allowed-Part-Of}) of the domain framework, enabling the reasoning agent to perform schema introspection before querying the graph.
    \begin{itemize}
    \item \texttt{id (ID):} A unique identifier for the taxonomic type.
    \item \texttt{label (string):} The human-readable label (e.g., ``Article,'' ``Constitution'').
    \item \texttt{urn (string)?:} (Optional) A persistent semantic identifier.
    \item \texttt{url (string)?:} (Optional) A dereferenceable URI providing Web access.
    \item \texttt{parentIds (list[ID])?:} (Optional) A list of IDs for broader superclasses.
    \item \texttt{childrenIds (list[ID])?:} (Optional) A list of IDs for narrower subclasses.
    \item \texttt{allowedPartsIds (list[ID])?:} (Optional) A list of \texttt{ItemType} IDs that are schema-permitted to act as immediate structural children.
    \item \texttt{metadata (JSON)?:} (Optional) A flexible JSON store for domain-specific drafting or schema rules and auxiliary heuristics, such as numbering patterns or canonical abbreviations.
    \end{itemize}
    
    \item[\texttt{Version}:] Represents a specific, time-bound snapshot of an \texttt{Item}. A \texttt{Version} object is the agent's key to the past, capturing the valid structural position and temporal state of an item at a specific point in time, while remaining cleanly decoupled from textual content, which is represented separately by \texttt{TextUnit}.
    \begin{itemize}
        \item \texttt{id (ID):} A unique identifier for this specific version.
        \item \texttt{itemId (ID):} The ID of the abstract \texttt{Item} it is a version of.
        \item \texttt{type (string):} A discriminator (e.g., \texttt{statutory}).
        \item \texttt{urn (string)?:} (Optional) A persistent semantic identifier.
        \item \texttt{url (string)?:} (Optional) A dereferenceable URI providing Web access.
        \item \texttt{validityInterval ([date, date]):} A tuple representing the start and end dates (RFC 3339) of its formal validity. The end date is optional (or null) to represent a currently-valid version.
        \item \texttt{applicabilityInterval ([date, date])?:} (Optional) A tuple defining material applicability, separating formal validity from concrete legal effects (e.g., \textit{vacatio legis}). When absent, implementations may treat applicability as coincident with \texttt{validityInterval}.
        \item \texttt{producedByActionId (ID):} The ID of the \texttt{Action} that created this version, enabling deterministic \textbf{backward traceability} (provenance).
        \item \texttt{terminatedByActionId (ID)?:} (Optional) The ID of the \texttt{Action} that ended this version. A null value indicates it is currently in force.
        \item \texttt{usedAsSourceInActionIds (list[ID])?:} (Optional) A list of \texttt{Action} IDs that used this specific version as the authorizing command to modify other entities in the graph. This enables deterministic \textbf{forward impact analysis}.
        \item \texttt{metadata (JSON)?:} (Optional) A flexible JSON store for structured properties of the version.
    \end{itemize}

    \item[\texttt{TextUnit}:] Represents the vectorizable informational container. It holds a textual or semantic aspect associated with a graph entity, acting as the interface between semantic discovery and deterministic graph traversal.
    \begin{itemize}
        \item \texttt{id (ID):} A unique identifier for this piece of text.
        \item \texttt{sourceType (string):} The polymorphic type of the target node (e.g., \texttt{Theme}, \texttt{Item}, \texttt{Version}, \texttt{Action}, \texttt{Relation}), enabling unambiguous structural resolution by the agent.
        \item \texttt{sourceId (ID):} The unique identifier of the node to which this text is attached.
        \item \texttt{language (string):} The standardized language code of the content (e.g., BCP 47 \texttt{pt-BR}).
        \item \texttt{aspect (string):} The semantic role of the text relative to its source node. Examples include the \texttt{canonical} text of a \texttt{Version}, \texttt{indexical} aliases (e.g., ``Clean Record Act''), \texttt{summary} descriptions, or \texttt{textual\_metadata} representing linearized structured attributes.
        \item \texttt{content (string):} The raw textual content to be consumed by the language model (and implicitly embedded in the vector database).
        \item \texttt{metadata (JSON)?:} (Optional) A JSON store preserving the original structured properties that were linearized into the \texttt{content}, enabling deterministic post-retrieval filtering in hybrid search strategies.
    \end{itemize}
    
    \item[\texttt{Action}:] Represents a reified state-transition event, acting as the mutation operator within the graph. An \texttt{Action} is not a generic event or actor-participation record; it is used only when an authority-bearing event produces, terminates, or modifies one or more \texttt{Version}s. In the legal instantiation, an \texttt{Action} may represent a promulgation, amendment, revocation, correction, or other legally operative event that changes the graph state. The model therefore connects the authorizing legal command to the specific topological transformations it enacts. It unifies macro-events (document promulgations) and micro-events (atomic granular changes) into a first-class entity. 
    \begin{itemize}
        \item \texttt{id (ID):} A unique identifier for the action.
        \item \texttt{type (string):} The semantic nature of the operation (e.g., \texttt{Promulgation}, \texttt{Amendment}, \texttt{Revocation}).
        \item \texttt{parentId (ID)?:} (Optional) The ID of the parent macro-action (e.g., the promulgating law) that contains this granular micro-action. A null value denotes a root macro-event.
        \item \texttt{eventTime (date):} The formal registration timestamp of the event (e.g., publication date in the official gazette).
        \item \texttt{validityInterval ([date, date]):} The formal command of validity dictated by this action, from which resulting \texttt{Version}s derive their temporal boundaries.
        \item \texttt{applicabilityInterval ([date, date])?:} (Optional) The formal command of material efficacy (e.g., to model \textit{vacatio legis}).
        \item \texttt{sourceVersionIds (list[ID]):} A list of IDs for the specific \texttt{Version}s of the legal text that \textbf{authorize} or command this change.
        \item \texttt{terminatesVersionIds (list[ID])?:} (Optional) A list of \texttt{Version} IDs whose validity is \textbf{terminated} by this action.
        \item \texttt{producesVersionIds (list[ID])?:} (Optional) A list of \texttt{Version} IDs that are \textbf{created} and injected into the graph by this action.
        \item \texttt{metadata (JSON)?:} (Optional) A flexible JSON store for structured properties of the action.
    \end{itemize}
    
    The optionality of \texttt{terminatesVersionIds} and \texttt{producesVersionIds} allows the same model to unambiguously distinguish event types: a \textbf{Creation} has only \texttt{producesVersionIds}; a pure \textbf{Revocation} has only \texttt{terminatesVersionIds}; a standard \textbf{Amendment} has both. We note that the reciprocal references between \texttt{Action} and \texttt{Version} (e.g., \texttt{Version.producedBy ActionId} and \texttt{Action.producesVersionIds}) describe the same underlying edge from opposite ends; both directions are maintained automatically by the ingestion pipeline and exposed by the API to enable traversal efficiency in either direction without requiring secondary lookups.

    \item[\texttt{Relation}:] Represents transversal semantic connections, such as citations, regulatory dependencies, implementation links, interpretive relations, or other typed cross-references. This enables the agent to perform ``Network Reasoning'' beyond the strict mereological hierarchy.
    \begin{itemize}
        \item \texttt{id (ID):} A unique identifier for the relation edge.
        \item \texttt{sourceId (ID)} and \texttt{sourceType (string):} The polymorphic origin of the relation (pointing to either an \texttt{Item} or a \texttt{Version}).
        \item \texttt{targetId (ID)} and \texttt{targetType (string):} The polymorphic destination of the relation.
        \item \texttt{predicate (string):} The semantic nature of the edge (e.g., \texttt{eli:cites}).
        \item \texttt{validityInterval ([date, date])?:} (Optional) The time interval during which this specific connection is legally valid.
        \item \texttt{metadata (JSON)?:} (Optional) A flexible JSON store for structured properties of the relation.
    \end{itemize}
\end{description}

\paragraph{Unified but Asymmetric Hierarchical Navigation.}
The API unifies several forms of hierarchy under a common navigation pattern: structural containment for \texttt{Item}s, taxonomic subsumption for \texttt{Theme}s and \texttt{ItemType}s, and temporal aggregation for \texttt{Version}s. The JSON representation is deliberately asymmetric. High-cardinality entities such as \texttt{Item}s expose only ascending pointers, such as \texttt{parentId}, while descendant discovery is delegated to explicit API primitives. Lower-cardinality entities such as \texttt{Theme}s and \texttt{ItemType}s may expose both parent and child pointers. This design avoids super-node payloads while preserving a uniform reasoning pattern for the agent: navigate upward to broaden context, downward to specialize it, and laterally through typed relations when needed.

\paragraph{Diachronic Efficiency through Structural Sharing.}
A major challenge in temporal graphs is avoiding combinatorial growth as entities change over time. The SAT-Graph data model uses structural sharing for \texttt{Version} entities. When an event modifies a granular component, new versions are instantiated only for the affected path; sibling components that were not modified retain their original pointers and are shared across reconstructions. This allows the graph to represent full historical states without materializing a complete copy of the entire document after each amendment. The result is a compact diachronic graph in which historical reconstruction depends primarily on the intrinsic size of the reconstructed structure rather than on the total number of past events.

\paragraph{Bi-Temporal Foundations.}
The model distinguishes valid time from transaction time~\cite{kulkarni2012temporal}. Valid time captures when a fact, version, or relation is considered operative in the represented domain. Transaction time captures when the system learned, recorded, or accepted that fact. This distinction is necessary for retroactive events, corrections, annulments, and other cases in which the present view of the past differs from the knowledge available at the past moment itself. The temporal primitives expose this distinction through parameters such as \texttt{at} and \texttt{observerTime}.

\paragraph{Vector Index Compaction for RAG Optimization.}
The separation between temporal topology and informational content also improves retrieval. Because textual content is stored in \texttt{TextUnit}s and linked to graph entities rather than blindly duplicated for every reconstructed historical state, unchanged provisions need not be re-indexed as near-identical vector entries. This reduces the risk of top-$k$ saturation by repeated copies of the same text and helps preserve retrieval diversity during semantic anchoring. The benefit is conditional on the indexing implementation, but the data model is designed to support compact, deduplicated vector indices.

\subsection{Canonical Primitive API Specification}
\label{subsec:api_spec}

The SAT-Graph API exposes a controlled vocabulary of primitives organized by reasoning intent. Each primitive is designed to be small enough to remain auditable and composable, but expressive enough to support complex execution plans when chained by an agent. The API therefore functions as a typed interface between probabilistic planning and deterministic graph execution.

The full API specification includes additional endpoints for convenience operations, batch retrieval, administrative tasks, and deployment-specific extensions. This paper focuses on the conceptual primitive layer and presents a representative core subset: the primitives needed to demonstrate how an agent moves from semantic discovery to deterministic temporal, structural, causal, and textual retrieval. The complete endpoint surface is maintained in the project repository.

The primitive layer is divided into functional groups: discovery and search, temporal resolution, structural navigation, causality and impact analysis, introspection, semantic traversal, and materialization/hydration. Discovery primitives may return ranked candidates or scored results because they operate at the boundary between language and graph identity. Primitives that operate over already resolved canonical identifiers are deterministic read operations, conditional on the current graph state. Table~\ref{tab:primitive_groups} summarizes the functional groups discussed in this paper.

\begin{table}[!htbp]
\centering
\footnotesize
\setlength{\tabcolsep}{4pt}
\renewcommand{\arraystretch}{0.92}
\caption{Functional primitive groups discussed in this paper. The table is representative rather than exhaustive; the complete endpoint surface is maintained in the project repository.}
\label{tab:primitive_groups}
\begin{tabular}{@{}p{0.18\textwidth} p{0.30\textwidth} p{0.42\textwidth}@{}}
\toprule
\textbf{Group} & \textbf{Representative primitives} & \textbf{Role in agentic reasoning} \\
\midrule

Discovery and search &
\texttt{resolveItemReference}, \texttt{searchTextUnits}, \texttt{searchItems} &
Translate natural-language references or semantic queries into candidate graph anchors. \\

Temporal resolution &
\texttt{getValidVersions}, \texttt{getApplicableVersions}, \texttt{getItemVersions} &
Resolve point-in-time state, version history, and material applicability after canonical anchoring. \\

Structural navigation &
\texttt{getItemHierarchy}, \texttt{getItemChildren}, \texttt{getItemAncestors} &
Reconstruct topological context and navigate part-whole or taxonomic structures. \\

Causality and impact &
\texttt{getItemHistory}, \texttt{getActionsBySource}, \texttt{queryActions} &
Trace provenance, amendments, revocations, and forward impact through reified events. \\

Introspection &
\texttt{getRootItemTypes}, \texttt{getRootThemes}, \texttt{getSupportedActionTypes} &
Allow the agent to discover valid schema vocabularies before formulating plans. \\

Semantic traversal &
\texttt{getRelations} &
Traverse typed cross-references such as citations, delegations, regulatory dependencies, or implementation links. \\

Hydration and materialization &
\texttt{getItemById}, \texttt{getVersionById}, \texttt{getBatchItems}, \texttt{getVersionTextUnits} &
Convert lightweight identifiers into full objects and textual evidence for synthesis. \\

\bottomrule
\end{tabular}
\end{table}

\paragraph{Architectural Rationale.}
The API deliberately encapsulates graph access through a procedural primitive surface rather than exposing SQL, SPARQL, or Cypher directly. This design responds to four risks: confabulation in generated queries, especially when temporal filters or recursive joins are required~\cite{pourreza2024dinsql}; the need to centralize non-trivial temporal resolution rather than reconstruct it in prompts; decoupling from the physical storage schema through a facade pattern~\cite{gamma1994design}; and reduction of the prompt-injection attack surface through a typed allowlist of read-only primitives~\cite{greshake2023more,perez2022ignore}.

\subsubsection{Discovery and Search Primitives}

Discovery primitives are the only primitives that intentionally operate under semantic uncertainty. They are used when the agent must translate a natural-language reference, thematic description, or content query into candidate graph anchors. Their role is not to produce final legal evidence, but to generate explicit hypotheses that can be verified through deterministic primitives.

This design separates two forms of uncertainty. \textit{Identity uncertainty} concerns which graph entity the user intended. It is handled by reference-resolution primitives that return ranked candidates. \textit{Relevance uncertainty} concerns how closely a textual unit matches a semantic query. It is handled by search primitives that return scored results. Both forms of uncertainty are surfaced to the agent rather than hidden inside a single retrieved context window.

\newpage
\vspace{0.3cm}
\noindent\textbf{\texttt{resolveItemReference(referenceText: string, contextId?: ID,}} \newline
\textbf{\texttt{at?: date, topK?: int) $\rightarrow$ list[ResolvedItemCandidate]}}

\begin{itemize}
    \item \textit{Description:} Performs time-aware entity linking. It maps a natural-language reference to a ranked list of candidate \texttt{Item} objects.
    \item \textit{Parameters:}
    \begin{itemize}
        \item \texttt{referenceText}: A reference such as ``Article~5 of the Constitution'' or a canonical URN.
        \item \texttt{contextId}: Optional structural context for resolving relative or ambiguous references.
        \item \texttt{at}: Optional reference time for resolving historically unstable labels, renumbered provisions, or time-dependent identifiers.
        \item \texttt{topK}: Optional maximum number of candidates.
    \end{itemize}
    \item \textit{Returns:} A ranked list of \texttt{ResolvedItemCandidate} objects, each containing an \texttt{Item} and a confidence score.
    \item \textit{Discussion:} The primitive does not force a single interpretation of an ambiguous reference. It exposes candidate identity and confidence to the agent. If confidence is below a configured threshold, the agent can request clarification or attempt further contextual grounding before invoking deterministic primitives.
\end{itemize}

\vspace{0.3cm}
\noindent\textbf{\texttt{searchTextUnits(itemIds?: list[ID], themeIds?: list[ID],}} \newline
\textbf{\texttt{itemTypeIds?: list[ID], metadataFilter?: MetadataFilter,}} \newline
\textbf{\texttt{at?: date, semanticQuery?: string, lexicalQuery?: string,}} \newline
\textbf{\texttt{language?: string, aspects?: list[string], topK?: int)}} \newline
\textbf{\texttt{$\rightarrow$ list[SearchedTextUnitResult]}}

\begin{itemize}
    \item \textit{Description:} Performs hybrid search over \texttt{TextUnit}s using semantic, lexical, structural, temporal, and metadata constraints. It returns scored textual candidates rather than final evidence.
    \item \textit{Parameters:}
    \begin{itemize}
        \item \texttt{itemIds}, \texttt{themeIds}, \texttt{itemTypeIds}: Optional structural or conceptual filters that constrain the search space before ranking.
        \item \texttt{metadataFilter}: Optional structured filter over \texttt{Item.metadata}.
        \item \texttt{at}: Optional valid-time constraint. When provided, only text units attached to versions valid at that instant are eligible. When omitted, the search may range across the full temporal index.
        \item \texttt{semanticQuery}, \texttt{lexicalQuery}: Optional dense and sparse query strings.
        \item \texttt{language}: Optional BCP~47 language filter.
        \item \texttt{aspects}: Optional filter over textual facets such as \texttt{canonical}, \texttt{summary}, \texttt{indexical}, or \texttt{textual\_metadata}.
        \item \texttt{topK}: Optional maximum number of results.
    \end{itemize}
    \item \textit{Returns:} A ranked list of \texttt{SearchedTextUnitResult} objects, each containing a \texttt{TextUnit}, score metadata, and its \texttt{sourceType}/\texttt{sourceId} anchor.
    \item \textit{Discussion:} This primitive implements a \textit{Search-as-Seed} pattern. Its results are not treated as final context to be injected directly into the LLM. Instead, each returned \texttt{TextUnit} provides an anchor from which the agent can launch deterministic verification: retrieve the owning version, check temporal validity, reconstruct hierarchy, inspect provenance, and only then decide whether the text should be used as evidence.
\end{itemize}

\paragraph{Other Discovery Primitives.}
The discovery layer also includes auxiliary primitives for non-textual anchoring. \texttt{resolveThemeReference} maps natural-language subject descriptions, such as ``Social Security'' or ``Environmental Law'', to ranked \texttt{Theme} candidates. \texttt{resolveItemTypeReference} maps references to structural classes, such as ``Article'' or ``Constitution'', to ranked \texttt{ItemType} candidates. \texttt{searchItems} performs hybrid discovery over stable \texttt{Item}s rather than over textual units. Whereas \texttt{searchTextUnits} retrieves candidate pieces of evidence, \texttt{searchItems} identifies durable graph entities that have, at some point in their history, been associated with a concept, expression, or metadata pattern.

\paragraph{Uncertainty-Aware Return Types.}
Discovery primitives make uncertainty explicit in their return types. \texttt{Resolved<Entity>Candidate} objects carry a candidate entity and a confidence score, representing identity uncertainty: \textit{which entity did the user mean?} \texttt{SearchedTextUnitResult} objects carry a retrieved text unit and a relevance score, representing relevance uncertainty: \textit{how closely does this text match the query?} The distinction matters because the downstream policies differ. Low identity confidence may require clarification before execution, while low relevance may simply cause the agent to discard a candidate during evidence selection.

\subsubsection{Temporal Resolution Primitives}
\label{subsubsec:temporal_primitives}

Temporal primitives resolve the state of an \texttt{Item} across valid time and transaction time. They are deterministic after anchoring: given a canonical \texttt{itemId}, temporal parameters, and a fixed graph state, they return the same \texttt{Version} objects. Their purpose is to remove semantic approximation from point-in-time retrieval and to make temporal assumptions explicit in the agent's execution plan.

\vspace{0.3cm}
\noindent\textbf{\texttt{getValidVersions(itemId: ID, at?: date, observerTime?: date)}} \newline
\textbf{\texttt{$\rightarrow$ list[Version]}}

\begin{itemize}
    \item \textit{Description:} Resolves the \texttt{Version} object or objects of an \texttt{Item} that are valid at a target time, evaluated from a specified observer time.
    \item \textit{Parameters:}
    \begin{itemize}
        \item \texttt{itemId}: The canonical ID of the abstract \texttt{Item}.
        \item \texttt{at}: Optional valid-time coordinate. It indicates the date-time for which validity is evaluated. Defaults to the current system time.
        \item \texttt{observerTime}: Optional transaction-time coordinate. It indicates the temporal perspective from which the graph is observed. Defaults to the current system time.
    \end{itemize}
    \item \textit{Returns:} A list of \texttt{Version} objects valid under the specified temporal coordinates.
    \item \textit{Discussion:} The return type is plural by design. In the simplest statutory case, the list typically contains one version. In more complex settings, multiple simultaneously valid versions may coexist, such as a statutory version and an interpretive or administrative overlay. The primitive does not decide which version controls the legal answer. It exposes the valid candidates and their metadata so that the agent can reason over type, provenance, authority, and textual aspect.
\end{itemize}

\vspace{0.3cm}
\noindent\textbf{\texttt{getItemVersions(itemId: ID, startAt?: date, endAt?: date)
\newline
$\rightarrow$ list[Version]}}

\begin{itemize}
    \item \textit{Description:} Retrieves the  chronological history of \texttt{Version} instances produced for a specific \texttt{Item}.
    \item \textit{Parameters:}
    \begin{itemize}
        \item \texttt{itemId}: The canonical ID of the \texttt{Item}.
        \item \texttt{startAt}: (Optional) A date-time filter for the beginning of the interval.
        \item \texttt{endAt}: (Optional) A date-time filter for the end of the interval.
    \end{itemize}
    \item \textit{Returns:} A chronologically ordered list of \texttt{Version} objects.
    \item \textit{Discussion:} Unlike \texttt{getValidVersions}, which answers ``which version was formally valid at time $t$?'', this primitive answers ``what is the evolutionary history of this item?''. The optional \texttt{startAt} and \texttt{endAt} parameters filter the results by versions whose \texttt{validityInterval} overlaps with the specified temporal window.
\end{itemize}

\paragraph{Validity versus Applicability.}
The temporal primitives distinguish formal validity from material applicability. \texttt{getValidVersions} resolves versions by their \texttt{validityInterval}; the returned \texttt{Version} objects also expose \texttt{applicabilityInterval}, allowing the agent to inspect whether a formally valid version was materially applicable at the queried time. The full API specification provides a dedicated primitive, \texttt{getApplicableVersions(itemId: ID, at?: date, observerTime?: date) $\rightarrow$ list[Version]}, which applies the same deterministic temporal semantics to \texttt{applicability Interval}. This keeps applicability filtering inside the deterministic layer rather than reimplemented in the LLM prompt, preserving Probability Isolation across both temporal dimensions.

\paragraph{Bi-temporal Semantics.}
The behavior of \texttt{observerTime} operationalizes the bi-temporal model introduced in \S\ref{subsec:data_models}: when set to a past date, the primitive returns the historical truth as observed at that date; when defaulted to the current time, it returns the present view of the past.

\paragraph{Batch Temporal Retrieval.}
The full specification includes batch validity retrieval, such as \texttt{getBatch ValidVersions}, allowing the agent to resolve the state of multiple items in one operation when reconstructing a large historical structure. These batch operations are convenience primitives over the same temporal semantics, not a separate reasoning model.

\subsubsection{Structural Navigation Primitives}
\label{subsubsec:structural_navigation}

Structural primitives allow the agent to navigate the topology of the graph after an entity has been anchored. In the legal instantiation, this includes mereological relations such as document $\rightarrow$ title $\rightarrow$ chapter $\rightarrow$ article $\rightarrow$ paragraph. More generally, it covers structural containment, taxonomic subsumption, and temporally anchored hierarchy reconstruction. These primitives support context reconstruction without relying on textual adjacency.

\vspace{0.3cm}
\noindent\textbf{\texttt{getItemHierarchy(itemId: ID, depth?: int) $\rightarrow$ list[ID]}}\\
Returns the IDs of structural descendants of an \texttt{Item}. The optional \texttt{depth} parameter bounds traversal: \texttt{depth=1} returns direct children, while omission or a negative value returns all descendants recursively. The primitive returns lightweight identifiers rather than hydrated objects because deep traversal may produce large subtrees. If the agent requires full metadata or text, it performs a subsequent batch hydration step.

\vspace{0.3cm}
\noindent\textbf{\texttt{getItemChildren(itemId: ID) $\rightarrow$ list[Item]}}\\
Returns the immediate structural children of an \texttt{Item}. Because the immediate branching factor is usually smaller than a full subtree, this primitive returns hydrated \texttt{Item} objects and is suitable for local structural exploration.

\vspace{0.3cm}
\noindent\textbf{\texttt{getItemAncestors(itemId: ID) $\rightarrow$ list[Item]}}\\
Returns the ordered ancestor chain of an \texttt{Item} up to the root. This primitive reconstructs the structural breadcrumb of a provision, allowing the agent to contextualize a retrieved fragment within its document, title, chapter, section, or other enclosing components.

\paragraph{Payload Asymmetry.}
The structural layer uses payload asymmetry to protect the agent's context window. Deep traversals return IDs; narrow traversals return hydrated objects. This distinction lets the agent explore large structures cheaply while reserving full object hydration for selected nodes.

The API also exposes temporally anchored structural primitives---\texttt{getVersionHierarchy} and \texttt{getVersionAncestors}---which operate over \texttt{Version} rather than \texttt{Item} objects. Because structural sharing may allow a version to participate in different historical reconstructions, bottom-up temporal navigation requires a temporal anchor. A primitive such as \texttt{getVersionAncestors(versionId: ID, at?: date)} resolves the ancestor chain active at the specified time, avoiding ambiguity among historical parent paths.

\paragraph{Taxonomy and Concept Navigation.}
The same navigation pattern applies to classification structures. Representative primitives include \texttt{getItemTypeHierarchy}, which expands structural type taxonomies; \texttt{getThemeHierarchy}, which expands conceptual taxonomies; and \texttt{getThemesForItems}, which maps retrieved items to their associated themes. These operations allow the agent to move between concrete graph entities and higher-level conceptual or schema categories.

\subsubsection{Causality and Impact Analysis Primitives}
\label{subsubsec:causality_primitives}

Causality primitives expose the event layer of the graph. They allow the agent to move from static state retrieval to causal explanation: which action produced a version, which action terminated it, which source entity authorized a state transition, and which targets were affected. In the legal instantiation, these primitives support amendment history, revocation tracing, provenance reconstruction, and forward impact analysis.

\vspace{0.3cm}
\noindent\textbf{\texttt{getItemHistory(itemId: ID, startTime?: date, endTime?: date,}} \newline
\textbf{\texttt{actionTypes?: list[string], granularity?: string) $\rightarrow$ list[Action]}}

\begin{itemize}
    \item \textit{Description:} Retrieves the chronologically ordered actions that affected a specific \texttt{Item}.
    \item \textit{Discussion:} This primitive supports retrospective causality. The optional \texttt{actionTypes} parameter filters the timeline by event type, such as amendment, revocation, or correction. The optional \texttt{granularity} parameter lets the agent distinguish macro-events, such as an amending act as a whole, from micro-events, such as the specific alteration of a provision.
\end{itemize}

\vspace{0.3cm}
\noindent\textbf{\texttt{getActionsBySource(sourceWorkId: ID, actionTypes?: list[string],}} \newline
\textbf{\texttt{granularity?: string) $\rightarrow$ list[Action]}}

\begin{itemize}
    \item \textit{Description:} Retrieves actions caused or authorized by a specified source entity.
    \item \textit{Discussion:} This primitive supports forward causality. It allows the agent to answer questions such as which provisions were altered by a constitutional amendment, regulation, or administrative act, provided that the source event is modeled as producing or terminating versions in the graph. The returned \texttt{Action} objects expose their targets through fields such as \texttt{producesVersionIds} and \texttt{terminatesVersionIds}.
\end{itemize}

\vspace{0.3cm}
\noindent\textbf{\texttt{queryActions(itemIds?: list[ID], producesVersionIds?: list[ID],}} \newline
\textbf{\texttt{actionTypes?: list[string], granularity?: string,
\newline
startTime?: date, endTime?: date)}} \textbf{\texttt{$\rightarrow$ list[Action]}}

\begin{itemize}
    \item \textit{Description:} Retrieves \texttt{Action} objects using structured filters over affected items, produced versions, event types, or temporal windows.
    \item \textit{Discussion:} This primitive is useful for impact analysis at scale. Instead of issuing separate history queries for many provisions, the agent can query the event layer once over a bounded set of items or versions. This shifts graph intersection and filtering to the backend while preserving an explicit action-level audit trail.
\end{itemize}

\paragraph{Dynamic Orchestration.}
The API does not impose a fixed causal workflow. The agent composes primitives according to the user's question. It may perform a provenance audit by moving from a version to the action that produced it, or a cascade analysis by moving from a source act to all actions and versions it affected. Because \texttt{Action} objects expose canonical identifiers for sources, produced versions, and terminated versions, the resulting causal path is inspectable and reproducible, conditional on the correctness of the underlying event model.

\subsubsection{Introspection and Self-Discovery Primitives}
\label{subsubsec:introspection_primitives}

A reasoning agent should not rely solely on a hardcoded vocabulary injected into its prompt. Ontologies evolve, jurisdictions use different document types, and implementations may expose different taxonomies. Introspection primitives allow the agent to query the schema-level boundaries of the graph before constructing a plan.

\vspace{0.3cm}
\noindent\textbf{\texttt{getRootItemTypes() $\rightarrow$ list[ItemType]}}\\
Returns the root nodes of the structural type taxonomy. The agent can use this primitive to begin schema exploration before applying type-constrained retrieval.

\vspace{0.3cm}
\noindent\textbf{\texttt{getRootThemes() $\rightarrow$ list[Theme]}}\\
Returns the root nodes of the conceptual taxonomy. This enables progressive thematic exploration without loading the full taxonomy into the prompt.

\vspace{0.3cm}
\noindent\textbf{\texttt{getSupportedActionTypes() $\rightarrow$ list[string]}}\\
Returns the canonical event types supported by the graph. The agent can use these values to formulate valid causal filters.

\vspace{0.3cm}
\noindent\textbf{\texttt{getAvailableLanguages() $\rightarrow$ list[string]}}\\
Returns the BCP~47 language codes available in the textual indices.

\paragraph{Agent Bootstrapping.}
Introspection primitives support lazy loading of schema knowledge. Instead of embedding a long and potentially obsolete manual of types, themes, and action labels into the system prompt, the agent can fetch the relevant vocabulary when needed. This reduces prompt size and lowers the risk of malformed filters caused by terminological confabulation.

\subsubsection{Semantic Traversal Primitives}
\label{subsubsec:semantic_traversal}

Structural navigation explores vertical hierarchy, and causality primitives explore temporal transformation. Many knowledge bases also contain horizontal semantic relations: citations, delegations, dependencies, implementation links, regulatory references, or interpretive relations. The API exposes these links through typed traversal primitives rather than through unconstrained graph queries.

\vspace{0.3cm}
\noindent\textbf{\texttt{getRelations(entityId?: ID, entityType?: string, predicate?: string,}} \newline
\textbf{\texttt{direction?: string, startTime?: date, endTime?: date)}} \newline
\textbf{\texttt{$\rightarrow$ list[Relation]}}

\begin{itemize}
    \item \textit{Description:} Retrieves \texttt{Relation} objects using an anchor-based traversal over typed graph edges.
    \item \textit{Parameters:}
    \begin{itemize}
        \item \texttt{entityId}: Optional anchor entity.
        \item \texttt{entityType}: Optional type of the anchor entity, such as \texttt{Item} or \texttt{Version}.
        \item \texttt{predicate}: Optional relation predicate, such as citation, implementation, regulation, or dependency.
        \item \texttt{direction}: Optional traversal direction: \texttt{forward}, \texttt{backward}, or \texttt{both}.
        \item \texttt{startTime}, \texttt{endTime}: Optional filters over the relation's own validity interval.
    \end{itemize}
    \item \textit{Discussion:} The primitive lets the agent discover outgoing or incoming semantic links without generating open-ended SPARQL or Cypher queries. When a relation has been extracted, validated, and stored in the graph, the agent can traverse it deterministically and inspect its metadata, predicate, temporal interval, and endpoints.
\end{itemize}

\paragraph{Cross-Reference Resolution.}
In legal corpora, cross-references often affect the interpretation or applicability of a provision. A statute may cite another statute, delegate regulation to an administrative act, or incorporate definitions from an external instrument. Through \texttt{getRelations}, the agent follows validated graph edges rather than relying on semantic similarity to guess the referenced context. This does not guarantee that every cross-reference has been extracted or modeled correctly, but it makes modeled references explicit, typed, and auditable.

\subsubsection{Materialization and Hydration Primitives}
\label{subsubsec:materialization_primitives}

Many navigation primitives return lightweight identifiers rather than full objects. Hydration primitives convert those identifiers into structured objects and textual evidence. This separation allows the agent to explore broadly, filter logically, and hydrate only the nodes needed for final synthesis.

\vspace{0.3cm}
\noindent\textbf{\texttt{getItemById(itemId: ID) $\rightarrow$ Item}}\\
Returns the hydrated \texttt{Item} object identified by \texttt{itemId}.

\vspace{0.3cm}
\noindent\textbf{\texttt{getVersionById(versionId: ID) $\rightarrow$ Version}}\\
Returns the hydrated \texttt{Version} object, including validity intervals and causal links.

\vspace{0.3cm}
\noindent\textbf{\texttt{getActionById(actionId: ID) $\rightarrow$ Action}}\\
Returns the hydrated \texttt{Action} object, including its source versions and the versions it produces or terminates.

\vspace{0.3cm}
\noindent\textbf{\texttt{getBatchItems(itemIds: list[ID]) $\rightarrow$ list[Item]}}\\
Returns multiple \texttt{Item} objects in a single operation.

\vspace{0.3cm}
\noindent\textbf{\texttt{getBatchVersions(versionIds: list[ID]) $\rightarrow$ list[Version]}}\\
Returns multiple \texttt{Version} objects in a single operation.

\vspace{0.3cm}
\noindent\textbf{\texttt{getVersionTextUnits(versionId: ID, language?: string, aspect?: string)}} \newline
\textbf{\texttt{$\rightarrow$ list[TextUnit]}}\\
Returns the \texttt{TextUnit} objects attached to a specific \texttt{Version}. Optional filters select language and textual aspect, such as \texttt{canonical}, \texttt{summary}, \texttt{indexical}, or another implementation-defined facet.

\paragraph{Discovery versus Hydration.}
The separation between discovery and hydration is central to context management. Broad exploration primitives may return only IDs or scored anchors. The agent then filters those candidates using temporal, structural, causal, or semantic criteria before hydrating the selected objects. This pattern reduces unnecessary payload transfer, mitigates context-window pressure, and avoids repeated single-object calls when batch hydration is available. The final primitive in this chain, \texttt{getVersionTextUnits}, supplies the textual evidence used for answer synthesis.

\subsection{The API as a Typed Operational Grammar for Agentic Reasoning}
\label{subsec:api_grammar}

The primitives specified in this section are not merely a collection of endpoints. They form a typed operational grammar for agentic reasoning over a temporal knowledge graph. The grammar covers the graph's main reasoning axes: semantic discovery, temporal resolution, structural navigation, causal explanation, semantic traversal, schema introspection, and evidence hydration.

By design, these primitives are low-level operations. A complex legal question is not expected to be answered by a single call. Instead, the agent's planning module decomposes the user's prompt into an execution plan whose nodes are primitive calls and whose edges are data dependencies. A candidate \texttt{Item} ID returned by a discovery primitive may feed a temporal primitive; a \texttt{Version} returned by temporal resolution may feed a text hydration primitive; an \texttt{Action} returned by provenance tracing may feed a forward-impact query.

The resulting plan is the main audit artifact of the architecture. It can be inspected before execution for validation, during execution for monitoring, and after execution for review. Probability remains present in the generation of the plan and in final prose synthesis, but the graph operations themselves are typed, logged, and reproducible after anchoring. This is the practical expression of the \textit{Probability Isolation} principle.

The following section illustrates this grammar through use cases aligned with the SAT-Graph RAG framework. These examples show how questions that are difficult for flat semantic RAG can be decomposed into explicit, verifiable sequences of primitive operations.

\subsection{Formal OpenAPI Specification}
\label{subsec:openapi}

The complete, machine-readable endpoint specification is provided as an OpenAPI~3.0 YAML document in the project repository.\footnote{Further examples, detailed execution plans, and the full OpenAPI specification are accessible at: \url{https://github.com/hmartim/sat-graph-api}.} Whereas this paper focuses on the conceptual primitive layer and a representative core subset, the OpenAPI document serves as the implementation-facing contract for the full endpoint surface. It enables developers and researchers to inspect request and response schemas, generate client SDKs, implement automated test suites, and deploy interactive documentation using standard OpenAPI tooling.

\section{Use Cases and Application}
\label{sec:use_cases}

This section illustrates how the Canonical Primitive API operationalizes the query patterns introduced in our prior SAT-Graph RAG work~\cite{demartim2025graphrag}. The numbering of the use cases deliberately mirrors that earlier taxonomy in order to show how the same high-level retrieval patterns can now be expressed as explicit primitive-based execution plans. Use Case~2-A is presented as a forward-causality variant of Use Case~2 rather than as an independent fourth pattern.

Each use case follows the same structure: a user's natural-language query is presented, the challenge it poses to standard RAG is identified, and the agent's execution plan---a sequence of typed primitive calls---is described. The goal is not to report empirical benchmark results, but to demonstrate the API's reasoning grammar: how a probabilistic agent moves from semantic anchoring to deterministic temporal, structural, causal, and textual retrieval.

\subsection{Use Case 1: Deterministic Point-in-Time Retrieval}

\begin{quote}
\textit{User Query: ``What was the text of the caput of Article 6 of the Brazilian Constitution on May 20th, 2001?''}
\end{quote}

\paragraph{Challenge.}
This question appears simple, but it requires temporal precision. A standard, temporally naive RAG system may retrieve a semantically similar version of Article~6 without verifying whether that version was in force on May~20, 2001. Even if temporal metadata were added to chunks as a filter, retrieval would still depend on selecting the correct textual fragment from a probabilistic ranking. The SAT-Graph API separates the task into two stages: first, resolve the abstract provision; second, deterministically retrieve the version valid at the target date.

\paragraph{Agent Execution Plan.}
The plan illustrates \textit{Probability Isolation}: the initial reference-resolution step is probabilistic, while subsequent temporal and textual retrieval steps operate over canonical identifiers.

\begin{enumerate}[noitemsep, topsep=0pt]
    \item \textbf{Ground the Reference (probabilistic):} The agent translates the natural-language reference into candidate \texttt{Item} IDs. The specificity of ``caput of Article~6 of the Brazilian Constitution'' should yield a high-confidence candidate, but the resolution remains auditable through the returned score.
    \begin{lstlisting}[style=tablecode]
resolveItemReference(
    referenceText="Article 6, caput of the Brazilian Constitution"
)
    \end{lstlisting}

    \item \textbf{Resolve the Valid Version (deterministic):} Using the selected \texttt{itemId}, the agent retrieves the version valid at the target date.
    \begin{lstlisting}[style=tablecode]
getValidVersions(
    itemId="...",
    at="2001-05-20"
)
    \end{lstlisting}

    \item \textbf{Hydrate the Textual Evidence (deterministic):} From the returned \texttt{Version} object, the agent retrieves the canonical textual unit in the appropriate language.
    \begin{lstlisting}[style=tablecode]
getVersionTextUnits(
    versionId="...",
    language="pt-BR",
    aspect="canonical"
)
    \end{lstlisting}
\end{enumerate}

\paragraph{Synthesized Outcome.}
The agent receives a \texttt{TextUnit} linked to a specific \texttt{Version}, including its validity metadata and canonical identifier. The LLM then synthesizes a concise answer from this verified evidence: ``On May~20, 2001, the caput of Article~6 read: `...'\,''. The answer remains inspectable because the execution log records the resolved provision, the temporal query, the selected version, and the textual evidence used for synthesis.

\subsection{Use Case 2: Causal Pinpointing and Version Comparison}

\begin{quote}
\textit{User Query: ``What were the exact textual differences in the caput of Article 6 of the Brazilian Constitution before and after the amendment that introduced the right to `housing' (direito à moradia)?''}
\end{quote}

\paragraph{Challenge.}
This query requires more than retrieving two text fragments. It asks the agent to connect a conceptual change---the introduction of the right to housing---to a specific legislative event and to compare the versions immediately before and after that event. A standard semantic RAG system can search for the word ``moradia'', but it does not by itself identify the causal action that produced the textual mutation or the precise predecessor version that was terminated.

\paragraph{Agent Execution Plan.}
The agent first resolves the provision, then inspects its version history, identifies the textual pivot, and traces the causal event that produced the new version.

\begin{enumerate}[noitemsep, topsep=0pt]
    \item \textbf{Ground the Reference:} The agent resolves the natural-language reference into candidate \texttt{Item} IDs.
    \begin{lstlisting}[style=tablecode]
resolveItemReference(
    referenceText="Article 6, caput of the Brazilian Constitution"
)
    \end{lstlisting}

    \item \textbf{Retrieve the Provision's Evolution:} The agent retrieves the chronological version history of the selected \texttt{Item}.
    \begin{lstlisting}[style=tablecode]
getItemVersions(
    itemId="..."
)
    \end{lstlisting}

    \item \textbf{Hydrate Texts and Identify the Pivot:} The agent retrieves the canonical text units for the returned versions and compares them chronologically. It observes the first version in which ``moradia'' appears, binds it as \texttt{versionAfter}, and binds the immediately preceding version as \texttt{versionBefore}.
    \begin{lstlisting}[style=tablecode]
[getVersionTextUnits(
    versionId=v.id,
    language="pt-BR",
    aspect="canonical"
) for v in versions]
    \end{lstlisting}

    \item \textbf{Hydrate the Causal Action:} The agent obtains the \texttt{producedByActionId} from \texttt{versionAfter} and hydrates the corresponding \texttt{Action}.
    \begin{lstlisting}[style=tablecode]
getActionById(
    actionId=versionAfter.producedByActionId
)
    \end{lstlisting}

    \item \textbf{Compute the Textual Difference:} The agent performs a deterministic string-level comparison between the canonical text of \texttt{versionBefore} and the canonical text of \texttt{versionAfter}, identifying the minimal insertion, deletion, or substitution that separates the two versions.
    
    \item \textbf{Trace the Authorizing Source:} The agent hydrates the source version referenced by the action and reconstructs its ancestor chain to identify the root amending norm. For clarity, this example assumes a single authorizing source; in multi-source actions, the agent would iterate over all entries of \texttt{sourceVersionIds}.
    \begin{lstlisting}[style=tablecode]
getVersionById(
    versionId=action.sourceVersionIds[0]
)

getItemAncestors(
    itemId=sourceVersion.itemId
)
    \end{lstlisting}
\end{enumerate}

\paragraph{Synthesized Outcome.}
The agent now has the before-version, the after-version, the canonical text of both versions, the minimal textual diff, and the causal action that produced the change. The LLM can synthesize an answer aligned with the user's request:

\begin{quote}
Before Constitutional Amendment No.~26/2000, the caput of Article~6 read: ``São direitos sociais a educação, a saúde, o trabalho, o lazer, a segurança, a previdência social, a proteção à maternidade e à infância, a assistência aos desamparados, na forma desta Constituição.''

After the amendment, it read: ``São direitos sociais a educação, a saúde, o trabalho, a moradia, o lazer, a segurança, a previdência social, a proteção à maternidade e à infância, a assistência aos desamparados, na forma desta Constituição.''

The exact textual difference was the insertion of ``a moradia,'' immediately after ``o trabalho,'' and before ``o lazer.'' No other words in the caput were changed by this amendment.
\end{quote}

The answer is not merely that the right to housing was introduced; it identifies the precise textual mutation and links it to the \texttt{Action} that produced \texttt{versionAfter}. Each claim can be traced to a \texttt{Version}, \texttt{TextUnit}, or \texttt{Action} object in the execution log.

\subsection{Use Case 2-A: Forward Causality and Cascade Impact Analysis}

As a direct counterpart to the retrospective tracking demonstrated in Use Case~2, this variant explores forward causality. Whereas Use Case~2 moves backward from a changed provision to the amending action that produced the textual mutation, Use Case~2-A moves forward from the amending norm to the provisions it altered, added, or revoked across the corpus.

\begin{quote}
\textit{User Query: ``Which specific provisions of the Brazilian Constitution were altered, added, or revoked by Constitutional Amendment No. 26 of February 14, 2000?''}
\end{quote}

\paragraph{Challenge.}
Broad impact analysis is difficult for systems that lack an event model. The provisions affected by an amendment may not contain the name or number of the amending norm in their current text. The relevant connection is therefore causal and structural rather than lexical. A semantic search for ``Constitutional Amendment No.~26'' may retrieve the amendment itself, but it does not necessarily identify every provision produced, terminated, or modified by the amendment.

\paragraph{Agent Execution Plan.}
The agent uses the graph's event layer to traverse from the source norm to the micro-actions it authorized.

\begin{enumerate}[noitemsep, topsep=0pt]
    \item \textbf{Ground the Source Reference:} The agent resolves the amending norm into a canonical \texttt{Item} ID.
    \begin{lstlisting}[style=tablecode]
resolveItemReference(
    referenceText="Constitutional Amendment No. 26 of 2000"
)
    \end{lstlisting}

    \item \textbf{Retrieve the Cascade Effect:} The agent uses the resolved source ID to query the event layer. The \texttt{granularity} parameter restricts the result to micro-actions representing concrete transformations applied to target provisions.
    \begin{lstlisting}[style=tablecode]
getActionsBySource(
    sourceWorkId="...",
    granularity="micro"
)
    \end{lstlisting}

    \item \textbf{Identify and Hydrate Targets:} The returned \texttt{Action} objects expose \texttt{producesVersionIds} and \texttt{terminatesVersionIds}. The agent collects these IDs and hydrates the corresponding versions.
    \begin{lstlisting}[style=tablecode]
getBatchVersions(
    versionIds=[...extracted from actions...]
)
    \end{lstlisting}
\end{enumerate}

\paragraph{Synthesized Outcome.}
The agent receives a bounded set of versions directly linked to the requested amendment through causal edges. The LLM synthesizes a response listing the affected provisions and the type of transformation applied to each one. The output is grounded not in keyword overlap, but in the modeled event layer: each affected provision can be traced back to a specific \texttt{Action} authorized by the source norm.

\subsection{Use Case 3: Thematic and Hierarchical Impact Analysis}

\begin{quote}
\textit{User Query: ``Summarize the evolution of all provisions related to the theme `Digital Security' since 2000.''}
\end{quote}

\paragraph{Challenge.}
This query combines thematic discovery, hierarchical expansion, and temporal analysis. A standard semantic RAG system may retrieve provisions that explicitly mention digital security, but it may miss provisions that fall under a relevant thematic container without repeating the same terminology. Conversely, it may retrieve semantically similar text outside the intended legal scope. The challenge is to construct a bounded set of relevant provisions before analyzing their historical evolution.

\paragraph{Agent Execution Plan.}
The agent performs a two-phase plan. First, it resolves the thematic and hierarchical scope. Second, it retrieves the historical actions affecting that bounded scope.

\textbf{Phase 1: Scope Resolution}
\begin{enumerate}[noitemsep, topsep=0pt]
    \item \textbf{Ground the Thematic Scope:} The agent resolves the natural-language theme into candidate \texttt{Theme} IDs and expands the selected theme to include descendant sub-themes.
    \begin{lstlisting}[style=tablecode]
resolveThemeReference(
    referenceText="Digital Security"
)

getThemeHierarchy(
    themeId="..."
)
    \end{lstlisting}

    \item \textbf{Find Anchor Items:} The agent searches for stable \texttt{Item}s explicitly associated with the selected theme or its descendants.
    \begin{lstlisting}[style=tablecode]
searchItems(
    themeIds=[...expanded theme IDs...]
)
    \end{lstlisting}

    \item \textbf{Expand the Hierarchical Scope:} For each anchor item, the agent retrieves structural descendants. It may prune redundant traversals when one anchor is already contained within another anchor's subtree.
    \begin{lstlisting}[style=tablecode]
[getItemHierarchy(
    itemId=anchorId,
    depth=-1
) for anchorId in anchorIds]
    \end{lstlisting}
\end{enumerate}

\textbf{Phase 2: Historical Analysis}
\begin{enumerate}[noitemsep, topsep=0pt, resume]
    \item \textbf{Retrieve Historical Events:} With the expanded set of \texttt{itemId}s, the agent queries the event layer for actions affecting those items since the target date.
    \begin{lstlisting}[style=tablecode]
queryActions(
    itemIds=[...expanded item IDs...],
    startTime="2000-01-01"
)
    \end{lstlisting}

    \item \textbf{Aggregate and Select Evidence:} The agent groups returned \texttt{Action} objects by type, date, source, or affected provision, and selectively hydrates the most relevant versions and text units for final synthesis.
\end{enumerate}

\paragraph{Synthesized Outcome.}
The LLM receives a structured set of actions and selected textual evidence bounded by thematic and hierarchical scope. It can synthesize a longitudinal summary such as: ``Since 2000, provisions associated with digital security have been affected by X actions involving Y structural components. A major change was Constitutional Amendment No.~115/2022, which introduced data protection as a fundamental right...'' The answer remains auditable because each summarized event traces back to a specific \texttt{Action}, affected \texttt{Item}, and, where needed, hydrated \texttt{TextUnit}.

\section{Conclusion}
\label{sec:conclusion}

This paper addressed the interaction problem that arises once legal knowledge has been represented as a structure-aware temporal graph. Prior work on SAT-Graph RAG established a verifiable substrate for modeling hierarchy, temporal versioning, applicability, and legislative causality in legal norms~\cite{demartim2025graphrag}. A representation layer alone, however, does not determine how an LLM-based reasoning agent should query that substrate. This work specified the interaction layer required to build what we call \textbf{deterministic legal agents}: LLM-based systems whose planning, semantic discovery, and narrative synthesis may remain probabilistic, but whose retrieval process becomes deterministic once natural-language references, thematic questions, or semantic queries have been mapped to one or more auditable graph anchors.

We proposed the \textbf{SAT-Graph API} as that interaction protocol: a canonical primitive interface for auditable reasoning over temporal knowledge graphs, developed and illustrated in the legal domain. The API exposes typed, atomic, and composable operations that allow an agent to resolve references, retrieve point-in-time versions, reconstruct hierarchical context, trace causal provenance, identify forward impact, traverse semantic relations, and hydrate textual evidence. The contribution of the paper is therefore not a new monolithic retrieval engine, but a formal interface through which agentic reasoning can be decomposed into explicit, inspectable primitive calls.

The central design principle is \textit{Probability Isolation}. The proposed architecture does not eliminate the probabilistic nature of LLM-based systems. Instead, it confines stochastic behavior to bounded stages: intent translation, initial semantic anchoring, and final narrative synthesis. Once a canonical identifier and relevant temporal parameters have been established, structural, temporal, causal, and hydration primitives execute as deterministic read operations over the graph, conditional on the correctness and completeness of the underlying graph state.

This design turns retrieval into an auditable execution process. The agent's answer is no longer grounded only in an opaque set of retrieved chunks. It is grounded in a sequence of operations: which entity was resolved, which version was selected, which action produced or terminated it, which structural context was reconstructed, and which text units were used for synthesis. The resulting execution log becomes the main audit artifact of the architecture.

\subsection{Implications for Legal AI Architecture}

The proposed API has several implications for the design of trustworthy legal AI systems.

\paragraph{Architectural Decoupling of Knowledge and Reasoning.}
The architecture separates three computational responsibilities. The \textbf{SAT-Graph} represents the curated legal substrate. The \textbf{Canonical Primitive API} provides the typed interaction protocol. The \textbf{LLM-based agent} performs planning, orchestration, and narrative synthesis. This separation allows each layer to evolve independently. The knowledge graph can be curated and validated without retraining the agent; the agent can improve as LLMs improve without changing the underlying legal representation; and the API can preserve a stable contract between the two.

\paragraph{Auditability as a Retrieval Property.}
In conventional RAG systems, auditability is often treated as a post-hoc feature: the system retrieves text fragments and later attempts to cite them. In the proposed architecture, auditability is part of retrieval itself. Each primitive call records a typed operation over a canonical object. This makes it possible to inspect not only which evidence was used, but how that evidence was obtained. The audit trail is therefore procedural rather than merely textual.

\paragraph{Generalizability Beyond the Legal Domain.}
Although the API is specified and demonstrated in legal settings, its abstractions are intentionally domain-portable. The core objects---\texttt{Item}, \texttt{Version}, \texttt{TextUnit}, \texttt{Action}, and \texttt{Relation}---can model other authority-governed domains in which entities have stable identities, textual realizations, version histories, and causal events. Potential applications include technical standards, regulatory compliance, institutional policies, engineering specifications, and other domains where historical correctness and provenance are central.

\paragraph{Architectural Boundaries.}
The API is not a substitute for legal reasoning. It is a controlled evidentiary interface. It answers questions about graph state, temporal validity, structural context, textual evidence, and causal provenance. Higher-level tasks---such as resolving normative conflicts, weighing authority, performing analogical reasoning, or producing a legal opinion---remain the responsibility of the reasoning agent and, in high-stakes contexts, of human professionals. The API improves the grounding of such reasoning; it does not eliminate the need for interpretation or review.

\subsection{Limitations}

Several limitations follow directly from the architecture.

\paragraph{Dependence on Graph Quality.}
The deterministic behavior of the API is conditional on the correctness, completeness, and granularity of the underlying knowledge graph. If a legislative event is missing, if a version interval is incorrectly modeled, or if a cross-reference has not been extracted, the API may execute correctly while returning an incomplete or misleading evidentiary basis. The architecture therefore shifts part of the reliability problem from retrieval to ingestion, curation, and validation of the graph.

\paragraph{Implementation Complexity.}
A production-grade implementation requires robust middleware, temporal indexing, event modeling, validation pipelines, and batch execution strategies. The API surface is intentionally simple from the agent's perspective, but server-side execution may involve non-trivial logic for validity, applicability, retroactivity, structural sharing, and causal traversal. This complexity is justified by the need to centralize temporal and structural logic in the API rather than reconstruct it through prompts, but it remains an engineering burden.

\paragraph{Semantic Anchoring Errors.}
Discovery primitives expose uncertainty rather than eliminating it. A natural-language reference may still be resolved to the wrong candidate, especially when the query is ambiguous, underspecified, or jurisdictionally unclear. Returning ranked candidates and confidence metadata mitigates this risk by making uncertainty visible, but it does not guarantee that the agent will always choose the correct anchor.

\paragraph{Semantic Planning Errors.}
The architecture also does not prevent all planning errors. An agent may invoke syntactically valid primitives that are logically inadequate for the user's question, such as querying the wrong instrument, omitting a relevant relation traversal, or failing to inspect a necessary authority layer. The execution log makes such errors auditable, but preventing them requires improvements in planning strategies, prompting, fine-tuning, tool-use policies, and human-in-the-loop safeguards.

\paragraph{Empirical Evaluation Gap.}
This paper provides a formal architectural specification rather than a quantitative benchmark. The use cases illustrate how the primitive grammar operationalizes temporal, structural, and causal retrieval patterns, but they do not measure performance against baselines. A full empirical evaluation remains necessary to quantify accuracy, temporal precision, provenance correctness, latency, robustness, and auditability relative to standard RAG and Graph RAG approaches.

\subsection{Future Directions}

The specification presented here opens several directions for future work.

\paragraph{Agent Planning and Text-to-Plan Learning.}
A central next step is improving the planning layer that maps natural-language legal questions to primitive execution plans. One practical direction is to develop a library of reusable skills: canonical primitive sequences for common legal research tasks such as point-in-time retrieval, amendment tracing, thematic impact analysis, and provenance reconstruction. A longer-term direction is supervised fine-tuning or preference optimization over synthetic and manually validated examples of query-to-plan mappings. Such models could learn to generate compact execution DAGs directly, reducing reliance on long prompts and improving latency.

\paragraph{Empirical Benchmarking.}
The architecture should be evaluated against baselines on tasks that require temporal and causal reasoning. Suitable metrics include version-selection accuracy, temporal correctness, causal-chain completeness, provenance-attribution accuracy, answer faithfulness, plan validity, and latency. Existing legal RAG benchmarks provide a starting point~\cite{pipitone2024legalbench}, but additional datasets are needed for authority-governed, diachronic tasks in which the correct answer depends on validity intervals, amendments, repeals, and provenance.

\paragraph{Dataset and Evaluation Corpus Construction.}
A natural empirical follow-up is the construction of a benchmark corpus over a legal instrument with a dense amendment history, such as the Brazilian Constitution or a heavily amended statutory code. The dataset should contain natural-language questions paired with gold execution traces, expected versions, relevant actions, textual evidence, and final answers. This would allow the primitive-based approach to be compared not only on generated answer quality, but also on the correctness of intermediate retrieval steps.

\paragraph{Integration with Interpretive and Jurisprudential Layers.}
The current paper focuses primarily on the API specification for temporal knowledge graphs, illustrated through legislative norms. Future work can further explore how the same primitives support interpretive overlays, judicial decisions, administrative interpretations, and authority conflicts. 

\paragraph{Runtime Monitoring and Verification.}
Finally, execution logs can be used not only for post-hoc explanation, but also for runtime verification. A system may reject plans that omit required temporal parameters, flag low-confidence anchors, require human approval for ambiguous identity resolution, or compare generated answers against retrieved evidence before returning them. These safeguards would turn the primitive execution trace into an active control mechanism for trustworthy legal AI.

\paragraph{Closing Remark.}
The SAT-Graph API defines a disciplined boundary between probabilistic language understanding and deterministic graph execution. By making this boundary explicit, typed, and auditable, the architecture provides a foundation for legal agents that can reason over temporal knowledge without hiding their retrieval process inside either vector similarity or unconstrained generated queries. The next step is empirical: to test this specification at scale and measure whether primitive-based retrieval delivers the expected gains in temporal correctness, provenance fidelity, and auditability.

\bibliographystyle{plain}
\bibliography{referencias}

\end{document}